\documentclass[jimaging,article,accept,pdftex,moreauthors]{Definitions/mdpi} 
\usepackage{epsfig}
\usepackage{graphicx}
\usepackage[ruled]{algorithm2e}
\usepackage{array}
\usepackage{amssymb} % Add this line to include the amssymb package
\usepackage{mwe}
\usepackage{subfig}
\firstpage{1} 
\pubvolume{1}
\issuenum{1}
\articlenumber{0}
\pubyear{2023}
\copyrightyear{2023}
\externaleditor{Academic Editor: Name}
\datereceived{23 January 2024} 
\daterevised{14 April 2024} % Comment out if no revised date
\dateaccepted{19 April 2024} 
\datepublished{ } 
\hreflink{https://doi.org/}
\Title{A Multi-Modal Foundation Model to Assist People with Blindness and Low Vision in Environmental Interaction}

\TitleCitation{A Multi-Modal Foundation Model to Assist People with Blindness and Low Vision in Environmental Interaction}

\Author{Yu Hao %MDPI: Please carefully check the accuracy of names and affiliations.
 $^{1,\dag}$, Fan Yang $^{1,\dag}$, Hao Huang $^{1}$, Shuaihang Yuan $^{1}$, Sundeep Rangan $^{1}$, John-Ross Rizzo $^{1,2 %MDPI: wrong affiliation citation number, please revise them in numerical order.
}$, Yao Wang $^{1}$ \mbox{and Yi Fang $^{1,3,}$*}}

\AuthorNames{Yu Hao, Fan Yang, Hao Huang, Shuaihang Yuan, Sundeep Rangan, John-Ross Rizzo , Yao Wang  and Yi Fang}

\AuthorCitation{Hao, Y.;  %MDPI: Please carefully check the accuracy of names.
Yang, F.; Huang, H.; Yuan, S.; Rangan, S.; Rizzo, J.-R.; Wang, Y.; Fang, Y.}
\address{%
$^{1}$ \quad  %MDPI: We suggested to add full name for it.
 Tandon School of Engineering, New York University, New York, %MDPI: Newly added  information. Please confirm.
 USA; yh3252@nyu.edu %MDPI: We added the email addresses here according to the submitting system. Please confirm.
 (Y.H.); fy2167@nyu.edu (F.Y.); hh1811@nyu.edu (H.H.); sy2366@nyu.edu (S.Y.); srangan@nyu.edu (S.R.);  johnross.rizzo@nyulangone.org (J.-R.R.); yw523@nyu.edu (Y.W.)\\
$^{2}$  \quad NYU %MDPI: Please confirm if this affiliation has the full information, if not, please revise.
 Langone Health, New York University, New York, USA %MDPI: Please add the city, state abbreviation and postal code (or ZIP code in the U.S.). If the postal code is not available, please provide the P.O. Box..
 \\
 $^{3}$  \quad New York University Abu Dhabi, %MDPI: For universities, the department/school/faculty/campus is required. Please try to provide this information.
 Abu Dhabi, %MDPI: Please add the postal code (or ZIP code in the U.S.). If the postal code is not available, please provide the P.O. Box.
 UAE %MDPI: revised the format, please confirm.
}

\corres{Correspondence: yfang@nyu.edu}

\firstnote{These authors contributed equally to this work.}
\abstract{People with blindness and low vision (pBLV) encounter substantial challenges when it comes to comprehensive scene recognition and precise object identification in unfamiliar environments. Additionally, due to the vision loss, pBLV have difficulty in accessing and identifying potential tripping hazards independently. Previous assistive technologies for the visually impaired often struggle in real-world scenarios due to the need for constant training and lack of robustness, which limits their effectiveness, especially in dynamic and unfamiliar environments, where accurate and efficient perception is crucial. Therefore, we frame our research question in this paper as: \textit{How %MDPI:  Please confirm if the italics should be retained. if not, please remove it. This also applies to the highlights below.
 can we assist pBLV in recognizing scenes, identifying objects, and detecting potential tripping hazards in unfamiliar environments, where existing assistive technologies often falter due to their lack of robustness?} We hypothesize that by leveraging large pretrained foundation models and prompt engineering, we can create a system that effectively addresses the challenges faced by pBLV in unfamiliar environments. Motivated by the prevalence of large pretrained foundation models, particularly in assistive robotics applications, due to their accurate perception and robust contextual understanding in real-world scenarios induced by extensive pretraining, we present a pioneering approach that leverages foundation models to enhance visual perception for pBLV, offering detailed and comprehensive descriptions of the surrounding environment and providing warnings about potential risks. Specifically, our method begins by leveraging a large-image tagging model (i.e., Recognize Anything Model (RAM)) to identify all common objects present in the captured images. The recognition results and user query are then integrated into a prompt, tailored specifically for pBLV, using prompt engineering. By combining the prompt and input image, a vision-language foundation model (i.e., InstructBLIP) generates detailed and comprehensive descriptions of the environment and identifies potential risks in the environment by analyzing environmental objects and scenic landmarks, relevant to the prompt. We evaluate our approach through experiments conducted on both indoor and outdoor datasets. Our results demonstrate that our method can recognize objects accurately and provide insightful descriptions and analysis of the environment for pBLV.}


\begin{document}

%%%%%%%%%%%%%%%%%%%%%%%%%%%%%%%%%%%%%%%%%%
%\setcounter{section}{-1} %% Remove this when starting to work on the template.
% \section{How to Use this Template}

% The template details the sections that can be used in a manuscript. Note that the order and names of article sections may differ from the requirements of the journal (e.g., the positioning of the Materials and Methods section). Please check the instructions on the authors' page of the journal to verify the correct order and names. For any questions, please contact the editorial office of the journal or support@mdpi.com. For LaTeX-related questions please contact latex@mdpi.com.%\endnote{This is an endnote.} % To use endnotes, please un-comment \printendnotes below (before References). Only journal Laws uses \footnote.

% The order of the section titles is different for some journals. Please refer to the "Instructions for Authors” on the journal homepage.

\section{Introduction}
The prevalence of visual impairment has reached alarming levels, affecting millions of individuals worldwide, as highlighted by recent estimates from the World Health Organization (WHO) \cite{pascolini2012global, hakobyan2013mobile}. The number of people experiencing moderate to severe visual impairment or complete blindness continues to rise steadily, with projections indicating a further surge in these numbers by 2050 \cite{world2014visual}. Visual impairment, whether partial or complete, presents significant challenges that profoundly impact various aspects of daily life for \mbox{pBLV \cite{massiceti2018stereosonic}.}  Among the critical tasks that pose difficulties for pVLB is visual search, which involves actively scanning the environment and locating a specific target among distracting \mbox{elements \cite{treisman1980feature}.} Even for individuals with normal vision, visual search can be demanding, especially in complex environments. However, for individuals with blindness or low vision, these challenges are further compounded \cite{mackeben2011target}. Those with peripheral vision loss, central vision loss, or hemi-field vision loss often struggle to pinpoint a particular location or search for objects due to reduced fields of view. They often require assistance to accurately identify the environment or locate objects of interest. Similarly, individuals experiencing blurred vision or nearsightedness encounter difficulties in identifying objects at varying distances. Color-deficient vision and low-contrast vision further exacerbate the challenges of distinguishing objects from the background when they share similar colors. In addition to understanding their surroundings and locating objects of interest, assessing potential risks and hazards within the visual environment becomes an intricate task, demanding a comprehensive analysis to ensure personal safety \cite{fernandes2019review}. 
% The magnitude of these challenges emphasizes the need for innovative solutions to enhance visual perception and empower visually impaired individuals in their daily lives.
Therefore, addressing the challenges faced by pBLV in environmental interaction holds profound significance due to the escalating prevalence of visual impairment globally, which substantially affects millions and is projected to increase further. These challenges, which include difficulties in visual search, object identification, and risk assessment in diverse environments, critically impact the independence, safety, and quality of daily life of pBLV. Innovatively enhancing visual perception for these individuals not only promises to mitigate these profound challenges, but also aims to empower them with greater autonomy and confidence in navigating their surroundings, thus fostering inclusivity and accessibility in society.

Current assistive technologies for pBLV \cite{kameswaran2018we, roentgen2008inventory, loomis1994personal} driven by computer vision approaches have led to the development of assistive systems that utilize object recognition \cite{redmon2018yolov3}, GPS navigation \cite{pace1995global}, and text-to-speech tools \cite{li2019neural}. While these technologies have provided valuable assistance to visually impaired individuals \cite{montello2009cognitive}, they still face certain challenges and limitations. One of the primary challenges with existing assistive technologies is their limited ability to provide comprehensive scene understanding and guidance to address the specific needs of visually impaired individuals. For instance, while many tools focus on specific functionalities, such as obstacle detection or route planning, they often fall short in delivering detailed descriptions and guidance based on user questions. The current solutions also lack the capability to generate contextually relevant information about objects, scenes, and potential risks in the environment, limiting an in-depth understanding of the environment for visually impaired individuals. Conversational search finds applications in various domains such as basic information retrieval, personal information search, product selection and travel planning, which facilitates information retrieval through \mbox{conversation \cite{radlinski2017theoretical}.} Additionally, these solutions, such as object detection \cite{redmon2018yolov3}, frequently encounter difficulties in real-world scenarios due to the need for constant training and adaptation. They exhibit a lack of robustness, which limits their effectiveness, particularly in dynamic and unfamiliar environments, in which accurate and efficient perception is crucial. This limitation hinders their ability to fully perceive and understand their surroundings, resulting in reduced independence and increased reliance on external assistance.

\begin{figure}[t] 
    
    \includegraphics[width=13.5cm]{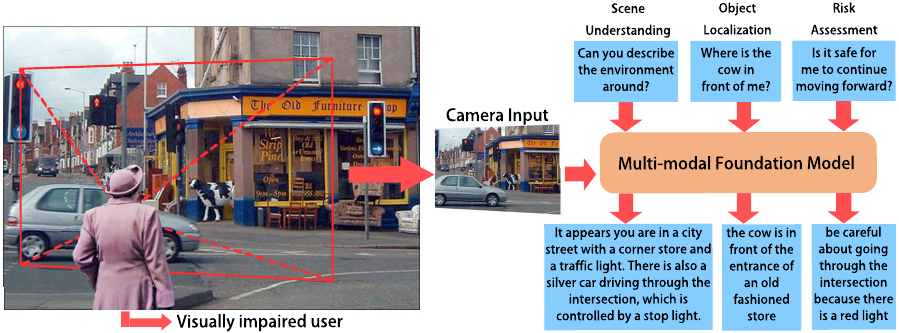}
    \caption{Multi-Modal %MDPI: Moved the figures after their first citation, please confirm the revisions in the whole paper.
 Foundation Model Sample Illustration.}
%\vspace{-1em}
    \label{fig1}
\end{figure}

In this paper, we aim to address the research questions of exploring whether large foundation models can address the limitations of current assistive technologies for pBLV by enhancing comprehensive scene understanding, providing contextually relevant information, and improving adaptability and robustness in dynamic environments. We posit that these large foundation models, through extensive pretraining, can significantly improve the functionality of assistive technologies by providing detailed guidance and accurate environmental perceptions, thus increasing the independence and safety of visually impaired individuals to navigate their surroundings. As shown in Figure \ref{fig1}, we present a novel approach named \textit{VisPercep} that leverages the advanced large vision-language model to enhance visual perception for individuals with blindness and low vision, including scene understanding, object localization, and risk assessment. Our work addresses the challenges faced by pBLV by providing them with detailed and comprehensive scene descriptions and risk guidance based on user questions, allowing an in-depth understanding of their surroundings, locating objects of interest, and identifying potential risks.

Our system includes three main modules, as illustrated in Figure \ref{fig2}: image tagging module, prompt engineering module, and vision-language module. The image tagging module, implemented using Recognize Anything Model (RAM) \cite{zhang2023recognize}, recognizes all objects in the captured image. We then integrate the recognized objects and user questions into a customized prompt designed for visually impaired individuals through prompt engineering. Finally, the vision-language model utilizes InstructBLIP \cite{li2023blip} to generate detailed and contextually relevant text, facilitating comprehensive scene understanding, object recognition, and risk assessment for visually impaired individuals. Our experiments demonstrate that our system can recognize objects of interest and provide detailed answers to user questions, significantly enhancing the visual understanding of surroundings.

Our contributions are summarized as follows:

\begin{enumerate}
    % \item We present a pioneering approach that integrates a multi-modal foundation model to enhance scene understanding for people with blindness and low vision (pBLV), offering detailed and comprehensive descriptions of the surrounding environment and providing warnings about potential risks.
    % \item We develop a voice-controlled system that facilitates language-guided question answering for pBLV. This innovative system integrates a large image tagging model with a large vision-language model.
    % \item We validate our approach through experiments on both indoor and outdoor datasets, demonstrating its ability to recognize objects accurately and provide insightful descriptions and analysis of the environment for pBLV.
    \item In response to the challenges faced by pBLV in achieving comprehensive scene recognition and identifying objects and hazards in unfamiliar environments, we introduce an innovative approach that leverages a multi-modal foundation model. This model is designed to significantly enhance environmental understanding by offering detailed and comprehensive descriptions of surroundings and alerting users to potential risks.
    \item To directly address the limitations of current assistive technologies, which often lack robustness and the capability to adapt to dynamic scenarios, we have developed a voice-controlled system. This system uniquely combines a large image tagging model with a vision-language foundation model, facilitating intuitive, language-guided question answering that caters specifically to the needs of pBLV.
    \item Our approach's effectiveness is validated through rigorous testing on both indoor and outdoor datasets. These experiments demonstrate the system's superior ability to accurately recognize objects and provide accurate descriptions and analyses of the environment, thereby directly addressing the core research problem of enhancing navigation and interaction for pBLV in diverse settings.
\end{enumerate}

In the following section of the article: (Section 2) Related Work: %MDPI: please confirm if these should be revised to ``Section 2'', ``Section 3''etc.
 Summarized existing auxiliary technologies in blind and low vision.
(Section 3) Materials and Methods: Proposed to leverage the power of a multi-modal foundation model that integrates image tagging and visual language models to provide detailed environmental descriptions and risk assessments. (Section 4) Experiments and Results: The accuracy and effectiveness of the model were validated through indoor and outdoor datasets, demonstrating its guidance ability for blind and visually impaired individuals. (Section 5) Conclusion: Summarized the contribution of the research, emphasizing the use of new methods to provide better guidance and enhance the independence and safety of visually impaired individuals. (Section 6) Limitations and Future Research: Explained the difficulties encountered by the model at the current stage from different aspects and future research directions.

\section{Related Works} 
Initial research has seen a growth in interest in the development of conversational search systems intended to support users in their information-seeking activities \cite{trippas2020towards}. This work has primarily focused on communication of information exclusively via spoken dialogue. While this is sufficient for simple question-type queries, it is an inefficient means of engagement for more complex or exploratory queries \cite{kaushik2020interface}. In the realm of information, conversational search is a relatively new trend \cite{kaushik2022exploratory}. Conversation is the natural mode for information exchange in daily life \cite{trippas2020towards}, and conversational approaches to information retrieval are gaining attention \cite{radlinski2017theoretical}. By integrating conversational search approaches with existing assistive technologies, there is potential to enhance the user experience and address the limitations of current systems, providing more natural and efficient interaction for individuals with visual disabilities.

In recent years, several assistive technologies and applications developed to support individuals with visual disabilities in understanding their environment and enhancing their scene understanding \cite{massiceti2018stereosonic, giudice2008blind, boldini2020piezoelectric}. Traditional tools such as white canes \cite{mcdaniel2008using} and guide dogs \cite{whitmarsh2005benefits} have long been used to aid in mobility and spatial awareness. Additionally, advancements in technology have led to the development of various assistive devices, including wearable cameras \cite{gupta2017indoor, rizzo2021covid, hao2022detect}, GPS navigation systems, and object recognition technologies 
 \cite{boldini2021inconspicuous}.

Wearable camera systems, such as the OrCam MyEye and Seeing AI \cite{granquist2021evaluation}, offer real-time text reading and text-to-speech capabilities to provide auditory feedback to individuals with visual disability. These systems assist in object identification, text reading, and facial recognition, enhancing their ability to interact with their surroundings. GPS navigation systems, such as BlindSquare \cite{kumar2022study} and Lazarillo \cite{cardosoaccessibility}, utilize location-based services to provide audio instructions and guidance for navigation in both indoor and outdoor environments.
   
Computer vision-based technologies have also been explored for scene understanding. These include object detection systems using deep learning models like YOLO \cite{redmon2016you} and Faster R-CNN \cite{ren2015faster}, which provide real-time identification of objects in the environment. Detect and Approach \cite{hao2022detect} proposes a real-time monocular-based navigation solution based on YOLO for pBLV. Additionally, vision-language models like VizWiz \cite{bigham2010vizwiz} and \mbox{SoundScape \cite{aletta2016soundscape}} incorporate natural language processing to describe visual scenes, answer questions, and provide context-aware information.

While these existing assistive technologies have made significant advancements, they still face limitations. Many systems provide partial solutions focused on specific functionalities such as object recognition or detection, but often fall short in delivering comprehensive scene understanding and detailed descriptions. Moreover, these technologies may lack the ability to provide guidance based on user questions, limiting their effectiveness in addressing the specific needs and queries of individuals with visual disability \cite{roentgen2008inventory}. Furthermore, these technologies often require multiple devices or interfaces, leading to complexity and decreased usability for individuals with visual disability \cite{giudice2008blind}. In contrast to these existing approaches, our proposed method offers a comprehensive and integrated solution. By combining advanced vision-language models, image tagging, and prompt engineering, our approach enhances scene understanding, provides real-time guidance, and offers context-aware prompts tailored specifically for individuals with visual disability. %Our method aims to overcome the limitations of existing assistive technologies and empower individuals with visual disability with improved navigation, enhanced object recognition, and comprehensive scene-understanding capabilities.
 
%%%%%%%%%%%%%%%%%%%%%%%%%%%%%%%%%%%%%%%%%%
\section{Materials and Methods}

\begin{figure}[t] 
    
    \includegraphics[width=13cm]{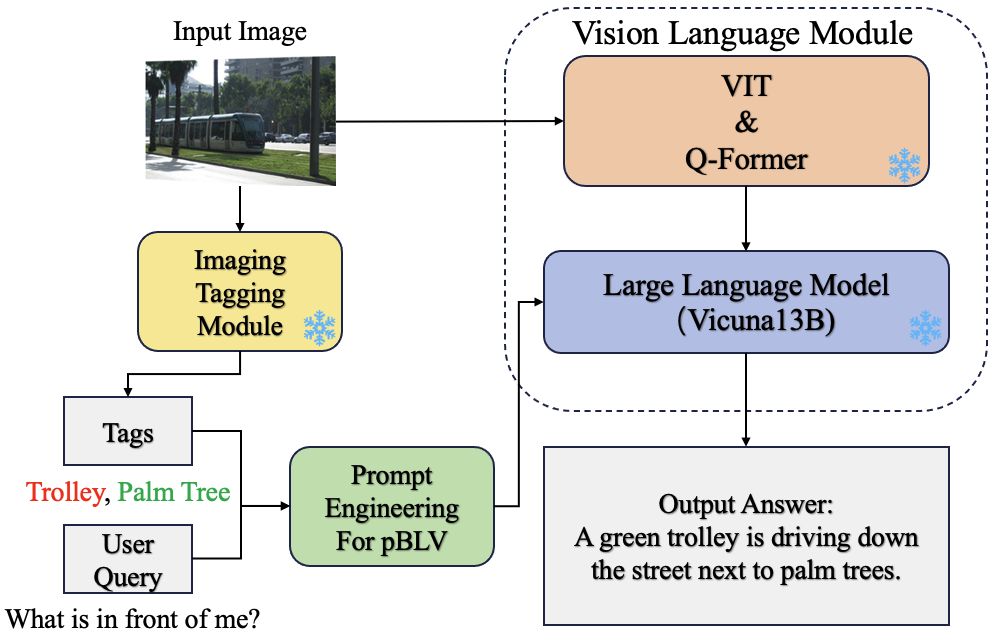}
    \caption{Method Structure Overview.} 
    %\vspace{-1em}
    \label{fig2}
\end{figure}

In this work, as shown in Figure \ref{fig1}, the proposed model leverages the advanced large vision-language model to assist environmental interaction for individuals with blindness and low vision including scene understanding, object localization, and risk assessment. Our system utilizes a smartphone to capture images and record user questions (left). Based on the input image and user question, our proposed model generates detailed and comprehensive scene descriptions and risk assessments (right). Moreover, the camera input image is from Visual7W dataset \cite{zhu2016cvpr}. Our work addresses the challenges faced by pBLV by providing them with detailed and comprehensive scene descriptions and risk guidance based on user questions, enabling an in-depth understanding of their surroundings, locating objects of interest, and identifying potential risks. 

Our system includes three main modules, as illustrated in Figure \ref{fig2}: image tagging module, prompt engineering module and vision-language module. Firstly, the image tagging module, implemented using Recognize Anything Model (RAM) \cite{zhang2023recognize}, identifies all common objects present in the captured image. Secondly, using prompt engineering, we integrate the recognized objects and user queries to create customized prompts tailored for individuals with visual disability. Lastly, the vision-language module which utilizes InstructBLIP \cite{li2023blip} generates detailed and contextually relevant output text, enabling comprehensive and precise scene understanding, object localization, and risk assessment for individuals with visual disability. [The input image is from Visual7W dataset \cite{zhu2016cvpr}.

Our method aims to overcome the limitations of existing assistive technologies and empower individuals with visual disability with improved guidance. In Section \ref{set1}, we introduce our image tagging module. Section \ref{set2} illustrates the prompt engineering tailored specifically for individuals with visual disability. We explain the large vision-language module in Section \ref{set3}.

%We introduce our large image tagging module in section \ref{set1}, enhanced object recognition in section \ref{set2}, and comprehensive scene-understanding capabilities in section \ref{set3}.

\subsection{Image Tagging Module}\label{set1}
As shown in the yellow box of Figure \ref{fig2}, the image tagging module is utilized to generate tags for each object present in the captured images, which is crucial as it provides a comprehensive understanding of the visual scene by accurately recognizing various objects. By incorporating the image tagging module, we obtain a catalog of objects present in the environment, facilitating a more precise and comprehensive environment description. We employ the Recognize Anything Model (RAM) \cite{zhang2023recognize} as our image tagging module, which has demonstrated the zero-shot ability to recognize any common category with \mbox{high accuracy.} 
 
%By incorporating RAM to generate tags prior to utilizing InstructBLIP, our method enhances the accuracy and effectiveness of the subsequent analysis performed by the vision language model. 
 
Specifically, the image tagging module begins with a pre-trained image encoder, which processes an input image $I$ and extracts high-level visual features $F$, formulated as: $F = Encoder(I)
$. These features capture important characteristics and representations of the objects in the image. After the initial feature extraction stage, an attention \mbox{mechanism \cite{liu2021swin}} is employed to focus on the most salient regions within the image. Represented mathematically as $A = Attention(F)$, this attention mechanism allows the model to pay more attention to relevant objects and suppress irrelevant ones. Thus, the image tagging module can generate accurate and informative tags for the recognized objects. The final stage involves mapping the extracted features to a set of object categories or tags by the image-tag
recognition decoder. This mapping, expressed as $T = Decoder(A)
$, is learned through a training process that leverages large-scale annotated datasets, ensuring the model's ability to generalize to various objects and scenes. The trained RAM model can then be applied to new images, accurately recognizing and generating tags for the objects present in the environment.

%By incorporating the object recognition module into our approach, we leverage its robust object recognition capabilities to generate tags for objects within the captured images. These tags provide valuable contextual information that enhances the analysis and understanding of the visual scene for the vision-language module that we will introduce in the following section.
%Our approach ensures a comprehensive and accurate vision-language framework for supporting individuals with visual disability in perceiving and navigating their environment effectively.

\subsection{Prompt Engineering for pBLV}\label{set2}
We incorporate prompt engineering, as shown in the green box of Figure \ref{fig2}, to create customized prompts tailored specifically for individuals with visual disability. This involves integrating the output of the image tagging module with user questions to form contextually relevant and informative prompts. Moreover, the use of prompt engineering eliminates the need for traditional machine learning approaches that require training models on labeled datasets, as prompt engineering focuses on generating effective prompts rather than optimizing model parameters.

The RAM generates a set of tags that represent the recognized objects within the captured images. We utilize these tags to enhance the final prompt. We include the prompt ``\textit{The image may contain elements of \{tags\}}'' to seamlessly integrate the object recognition results into a prompt. By incorporating these recognized object tags into the prompt, we ensure that the vision-language module receives specific and accurate information about the objects in their surroundings. This approach significantly enhances the understanding and awareness of the visual environment for the users.
    
Furthermore, we consider user questions as vital input for prompt engineering. By incorporating user questions into the prompts, we address the individual's specific needs for environmental understanding and ensure that the prompts are highly relevant to their current situation. This personalized approach allows individuals with visual disability to obtain the targeted information about their environment and the objects of interest. For example, in the case of risk assessment, we employ a specific prompt that guides the model to act as an assistant for individuals with visual disability, providing comprehensive analysis. %The prompt we use is ``I am visually disabled. You are a helper. Don't mention that I am visually disabled to offend me. [user\_query]. Your answer should be like ``Be careful about ... because ..."". 
The prompt we use is ``\textit{I am visually disabled. You are an assistant for individuals with visual disability. Your role is to provide helpful information and assistance based on my query. Your task is to provide a clear and concise response that addresses my needs effectively. Don't mention that I am visually disabled to offend me. Now, please answer my questions: [user\_query]. Your answer should be like a daily conversation with me.}'' where \textit{[user\_query]} is the user question. This prompt enables the model to deliver detailed and accurate explanations regarding potential risks, ensuring that the information is communicated in a respectful and \mbox{informative manner.}

\subsection{Vision-Language Module}\label{set3}
To generate output text based on the prompts obtained by the prompt-engineering module, we employ InstructBLIP \cite{li2023blip}, a powerful large vision language model for comprehensive scene understanding and analysis, as shown in the right blue box \mbox{of Figure \ref{fig2}.} %InstructBLIP utilizes a Query Transformer, or Q-Former, to extract visual features from a frozen image encoder, namely Vision Transformer (VIT). The Q-Former is responsible for processing the generated prompt tokens and extracting meaningful visual information from the input image.

Specifically, InstructBLIP begins by encoding the input image $I$ using the frozen Vision Transformer (VIT) \cite{dosovitskiy2020image}, which captures a high-level feature embedding $V$ of the image, represented as $V = VIT(I)$. %The input prompt is also encoded as the token by the tokenizer. 
%Then, the encoded prompt tokens and visual features are fed into the Q-Former \cite{li2023blip} together to generate contextualized image tokens through cross-attention.
The Q-Former \cite{li2023blip} in InstructBLIP, distinct from conventional models, employs learnable query embedding $Q$ and the image embedding $V$ from VIT for processing. This is formulated as $C = Q-Former(Q, V)$, where cross-attention is applied to generate contextualized soft image embedding $C$ \cite{li2023blip}.
The input prompt is also encoded as the high-dimensional prompt embedding $P$ by the tokenizer. 
%Then, a linear projection layer is employed to convert the image embeddings into the tokens that Large Language Model (LLM) can understand. 
%We further utilize a LLM, i.e., Vicuna-13B \cite{chiang2023vicuna}, to generate the final output text. 
The LLM incorporates both the image embedding and the prompt embedding from the user question to generate rich and comprehensive textual descriptions. 
Specifically, given the output of Q-Former as soft image embedding $C$ and prompt embedding $P$, the goal is to compute the probability of creating the final answer \(A\) with a length of \(N\) through the transformer model \(p_\theta\). The mathematical expression for this process is given by the equation \cite{zhang2023multimodal}:
\begin{equation}
p(A|V, P) = \prod_{i=1}^N p_\theta(A_i|C, P, A_{<i}) 
\end{equation}

This equation captures the sequential nature of language generation, where each embedding in the answer is dependent on the preceding embeddings, as well as the visual and prompt embeddings. This probabilistic approach ensures that the generated text is not only accurate but also contextually coherent. We demonstrate the algorithm of our proposed model in Algorithm \ref{alg:two}.
\vspace{12pt}

\SetKwComment{Comment}{/* }{ */}
\begin{algorithm}[H]
\caption{Algorithm %MDPI: Moved this after the first citation, please confirm.
 of Multi-modal Foundation Model.}\label{alg:two}
\KwIn{Image: The captured image  \\
\Indp      UserQuery: The user question}
\KwOut{OutputText: The generated output text }
\textbf{Step 1:} Predict Tags \\
\Indp
Image $\longrightarrow$ Image Tagging Module $\longrightarrow$ Tags \\
\Indm
\textbf{Step 2:} Prompt Engineering for pBLV \\
\Indp
Tags + UserQuery $\longrightarrow$ Prompt Engineering for pBLV $\longrightarrow$ Prompt \\
\Indm
\textbf{Step 3:} Generate OutputText \\
\Indp
Image + Prompt $\longrightarrow$ Vision-Language Module $\longrightarrow$ OutputText \\
\Indm
\end{algorithm}

% \begin{quote}
% This is an example of a quote.
% \end{quote}

%%%%%%%%%%%%%%%%%%%%%%%%%%%%%%%%%%%%%%%%%%
\section{Experiments}
\subsection{Implementation Details}
\begin{figure}[H]
    %\centering
    \includegraphics[width=5cm]{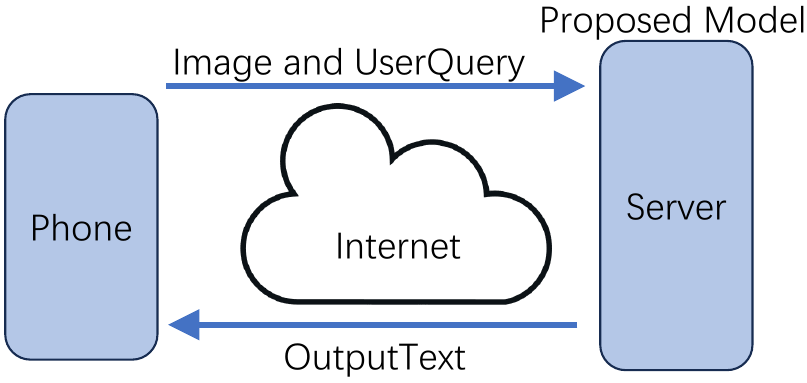}
    \caption{Client-server architecture. 
}
    %\vspace{-1em}
    \label{client-server}
\end{figure}

Our system leverages the capabilities of a smartphone, employing a monocular phone camera to capture images and the phone's microphone to receive user voice questions, creating a seamless interaction between the user and the system as shown in Figure %MDPI: wrong figure citation number, the previous one is Figure 2, please revise all of the citations in numerical order and add the figures after the first citation in the paper.
 \ref{client-server}. The image and voice input are then transferred to our server, where the processing and generation of comprehensive descriptions take place. To convert the user's voice question into text for further processing, we employ Whisper \cite{radford2023robust}, a powerful speech recognition system. This technology accurately transcribes the user's voice question into a textual form, enabling seamless integration with our vision-language model. After the input text is obtained, our system processes the image and text to generate detailed and contextually relevant output descriptions. The system selects the corresponding image frame once the user question is detected, ensuring accurate and timely responses. Here we utilize a LLM, i.e., Vicuna-13B \cite{chiang2023vicuna} ((Model settings include generating sentences with lengths ranging from 1 to 200 using beam search with a width of 5, applying a length penalty of 1, repetition penalty of 3, and temperature of 1), to generate the final output text. The output text is then transformed into audio format to provide a more accessible experience for individuals with visual disability. For text-to-speech conversion, we utilize the robust system \mbox{Azure \cite{li2019neural}.} This allows us to transform the output text into clear and natural-sounding audio. The synthesized audio is then sent from the server to the user's phone, enabling real-time delivery of the assist environmental interaction information. By implementing this client-server architecture and incorporating speech recognition and synthesis technologies, our system facilitates seamless interaction between the user and our system. %\blue{(Add a figure of the client-server architecture to summarize this section.)}

%Tests
\subsection{Tests on  Visual7W Dataset}      
%\subsection{Dataset Preparation}

We evaluate our proposed approach to the Visual7W dataset \cite{zhu2016cvpr}. Compared with previous studies that solely relied on textual answers, Visual7W introduces a novel form of question answering that includes visual answers \cite{zhu2016cvpr}. This is achieved by establishing a semantic connection through object-level grounding between textual descriptions and image regions \cite{zhu2016cvpr}. We notice that there are strong connections between objects in images, both in terms of spatial location and meaning of existence. To test our model in assisting people with visual disability, we selected some images from specific perspectives in this dataset. From these perspectives, pBLV often require additional assistance to better understand the current environment. In order to better adapt to the needs, we have set this task into three categories: Scene Understanding, Object Localization and Risk Assessment.

\subsubsection{Qualitative Performance Analysis for pBLV}

\begin{figure}[h] 
    
    \includegraphics[width=13.5cm]{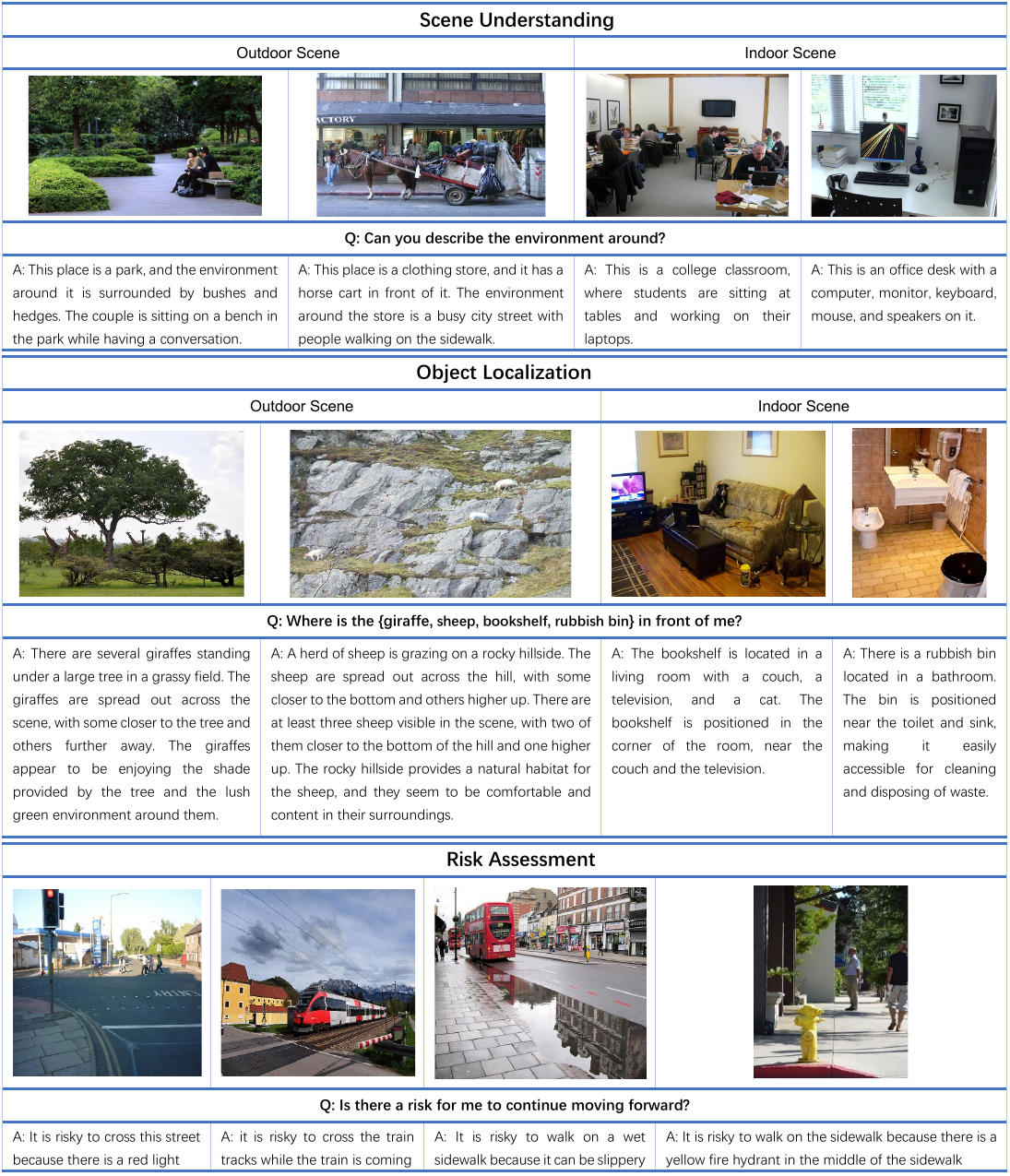}
    %\vspace{-0.5em}
    \caption{Examples of scene understanding (top), object localization (middle), and risk assessment (bottom) on Visual7W dataset.}
    %\vspace{-1em}
    \label{exp_1}
\end{figure}

\textbf{Scene Understanding: %MDPI: Please check if the bold should be retained? If not, please remove it. This also applies to the highlights below.
}We evaluate the effectiveness of our approach on outdoor and indoor scene understanding. Sample results are shown at the top of Figure \ref{exp_1}. In our experiment, the user's input is ``\textit{Can you describe the environment around?}''.
For both indoor and outdoor examples, it is evident that the model's output provides a comprehensive and accurate description of the object composition in the environment depicted in the image. The answer first summarizes names associated with the current place and then gives a specific description of objects and characters in the scene and what is happening at \mbox{this moment.}

\textbf{Object Localization:} We evaluate the effectiveness of our approach in addressing object recognition challenges, as demonstrated in the middle of Figure \ref{exp_1}.
The user question for this task is ``\textit{Where is the \{giraffe, sheep, bookshelf, rubbish bin\} in the image?}'', where ``\textit{\{ \}}'' is what the user wants to find out.

In the outdoor scene, the left image is focused on the giraffe. From the answer, we can see that the results are very detailed, not only describing the location of the giraffe on the grass and under the trees, but also providing contextual information ``\textit{The giraffes appear to be enjoying the shade provided by the tree and the lush green environment around them.}'' for users to better understand the capture environmental images.

% We also consider that finding objects is difficult for pBLV, especially in complex scenes where objects are numerous and messy. It is of vital importance for models to accurately determine the location of a specific objects and provide a description of the object's position. The instruction text for this task is “Describe the specific location of the {object} in the image using with as much detail as possible.”, i.e., where is the “{object}” that the user wants to recognize. More significantly, when describing the position of an object, it is necessary to describe its relative position in comparison to other objects, which will help people with visual disability better spatial understanding.

% \subsubsection{Outdoor Scene}

\textbf{Risk Assessment:} As shown at the bottom of Figure \ref{exp_1}, our model provides safety tips for people with visual disability to help them identify and deal with potential risks according to the current environment. The question is ``\textit{Is there a risk for me to continue moving forward?}''.

The first picture depicts a scene where a pedestrian crossing has a red light. The model can provide feedback to the user regarding the risk of crossing the street when the traffic signal is red. In the second scene, a train is approaching, which can be extremely dangerous if proper precautions are not taken. The model can send an alert that it is risky to cross the railway at the current time. It demonstrates that our model can effectively analyze risks and provide necessary alerts for pBLV.

\subsubsection{Quantitative Analysis of Inference Time and Helpfulness Scoring for pBLV}

In this section, we conduct a statistical quantitative analysis to evaluate our model's performance across three types of tasks. Specifically, we tested the inference time (in seconds) of the image tagging module and the vision-language module. As indicated \mbox{in Table \ref{tab:3task},} inference time testing was performed using an NVIDIA RTX A6000 GPU(48G). In our vision-language model, we utilize a byte-stream transmission approach during answer generation. This method allows users to receive the initial part of the answer even while the model is still generating subsequent segments. Consequently, we measure the inference time when the vision-language module generates the first token. The results demonstrate that our model can respond swiftly to user queries, with the total inference time (Image Tagging time plus Vision-language Inference time) for Object Localization being the fastest, at less than 0.3 s. These times hold potential for further reduction through software optimization.

Furthermore, it's important to assess how well the model's generated answers relate to and assist pBLV. Therefore, we employ a scoring system ranging from 0 to 10, where higher scores indicate greater relevance and helpfulness for pBLV. To determine these scores, we manually counted the number and names of important objects in each scene and compared them with the answers generated by the model. A point was deducted for each less important object identified in the generated answer. The final scores across all three tasks—Scene Understanding, Object Localization, and Risk Assessment—are reported in the table. Notably, Risk Assessment received the highest average score of 9.4, underscoring the model's effectiveness in providing relevant and helpful information to pBLV.

% \begin{table}[!ht]
% \centering
% \scalebox{1.0}{
% \begin{tabular}{@{}ccc@{}}
% \hline
% \multicolumn{1}{c}{Tasks} & \multicolumn{1}{c}{Inference time (Image Tagging / Vision-language)} & \multicolumn{1}{c}{Score of 10} \\ \hline
% Scene Understanding & 0.0419 / 0.3675  & 8.85 \\
% Object Localization & 0.0399 / 0.2359  & 8.6 \\
% Risk Assessment & 0.0356 / 0.2406  & 9.4 \\ \hline
% \end{tabular}}
% \caption{Quantitative Results of Inference time and Scoring on the Visual7W Dataset}
% \label{tab:3task}
% \end{table}

\begin{table}[H]
\caption{Quantitative Results of Inference time and Scoring on the Visual7W Dataset}
\label{tab:3task}

\newcolumntype{C}{>{\centering\arraybackslash}X}
\begin{tabularx}{\textwidth}{CCCC}
\toprule
\multicolumn{1}{l}{} & \multicolumn{2}{c}{\textbf{Inference time (seconds)}} & \multicolumn{1}{l}{} \\
\textbf{Tasks} & \textbf{Image Tagging} & \textbf{Vision-Language Inference} & \textbf{Score of 10} \\ \midrule
Scene Understanding & 0.0419 & 0.3675 & 8.85 \\
Object Localization & 0.0399 & 0.2359 & 8.60 \\
Risk Assessment & 0.0356 & 0.2406 & 9.40 \\ \bottomrule
\end{tabularx}

\end{table}

\subsubsection{Ablation Study}
\begin{figure}[h] 
    \centering
     \includegraphics[width=13.5cm]{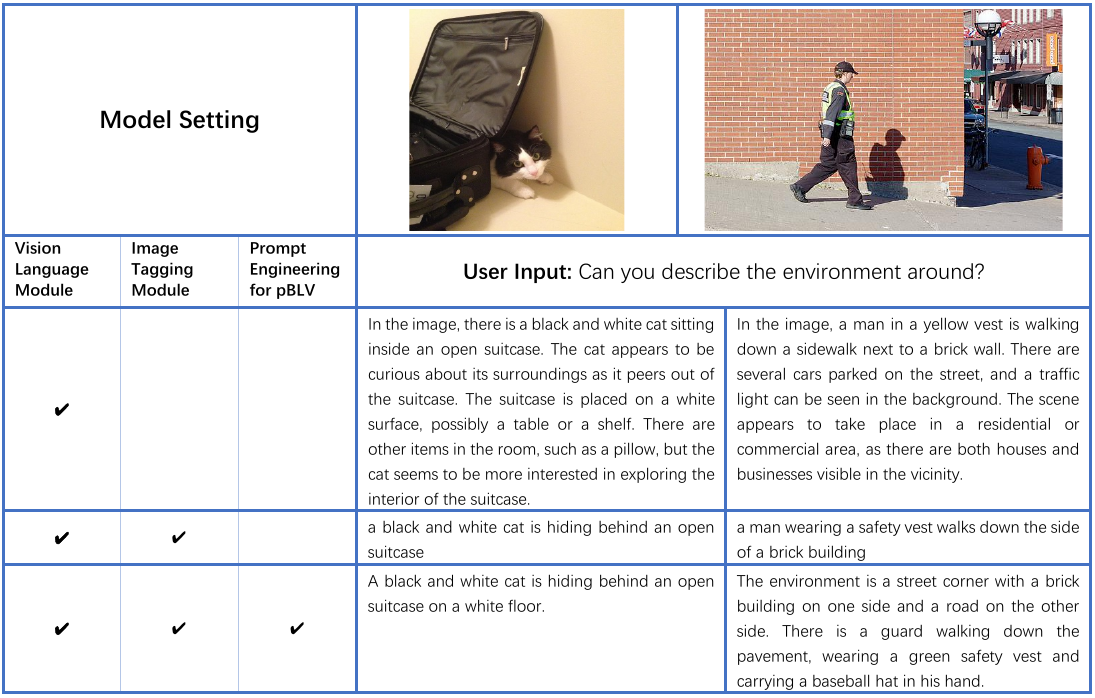}
     %\vspace{-0.5em}
    \caption{Ablation study with different model settings on Visual7W dataset.}
    %\vspace{-1em}
    \label{fig_ablsty}
\end{figure}

We conduct ablation studies to verify the effectiveness of the individual module in our model. The experimental settings are listed in Figure %MDPI: Wrong citation number (Figure 2, Figure 4, Figure 5, Figure 6, Figure 3), please revise it in numerical order.
 \ref{fig_ablsty} where ``\checkmark''
denotes the module is enabled. In the first experimental setting, we only utilize the vision-language module, which directly sends user questions and images to InstructBLIP. In the second experimental setting, we employ the image tagging module to generate tags for the input image, which are then integrated into the user question. Then, both the modified question and the input image are fed into the vision-language module. In the third experimental setting, we employ prompt engineering specifically designed for individuals with visual disability to further refine the prompt by incorporating the generated tags and user questions. 

As shown in Figure \ref{fig_ablsty}, for the left scene, when only using the vision-language module, the model provides some answers that do not match the facts, such as ``\textit{cat sitting inside an open suitcase}'', ``\textit{possibly a table or a shelf}'' and ``\textit{such as a pillow}''. These are more likely to be inferred from a large language model due to model's bias (learned from the training data) than what actually exists in the image. After combining with the image tagging module, the model dropped answers that do not match the facts in the image and the generated answer correctly describes the current scenario. The pipeline should describe the modified user questions after integrating them with the tags. Furthermore, if prompt engineering for pBLV is applied, answers become more precise e.g., it also accurately describes the location of the cat. Here you should describe what is the final prompt generated.

In the case of the right scene, the model that only uses the vision-language module does provide a detailed description of the scene, but there are still errors. The description ``\textit{There are several cars parked on the street, and a traffic light can be seen in the background''.} is inconsistent with the facts shown in the image, as there is no traffic light and only a white street light and an orange fire hydrant. When adding the image tagging module, the model gives a more factual description but lacks details. Again you should provide the new user question based on the tags.
%more examples
In contrast, Prompt Engineering for pBLV makes the answer more precise and detailed. Again you should provide the new user question after prompt engineering.

\begin{figure}[h]
    %\centering
    \includegraphics[width=8cm]{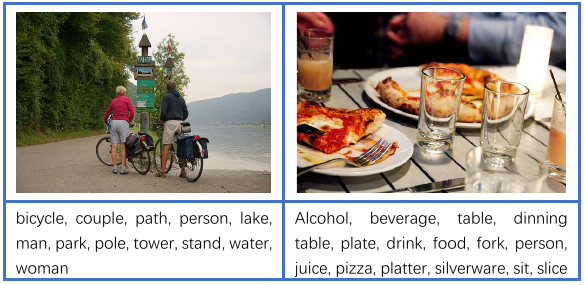}
    \caption{Random %MDPI: Figure 3 is not inserted after where it's first citation. Please revise the citation order and move it after the first citation.
 selected results of image tagging module on Visual7W dataset.}
    %\vspace{-1em}
    \label{ram}
\end{figure}

The example in Figure \ref{fig_ablsty} demonstrates that  the integration of the vision-language module, image tagging module, and prompt engineering yield the most accurate and detailed descriptions. In Figure %MDPI: Wrong citation number, please revise it in numerical order.
 \ref{ram}, we further present some randomly selected results of the image tagging module. As depicted in the figure, the module successfully recognizes common objects within the images, demonstrating its ability to provide a comprehensive understanding of the visual scene by accurately identifying various objects.

\vspace{-1em}

\subsection{Tests on VizWiz Dataset}
%先介绍VizWiz数据集，比如是vqa啊，有多少多少数据；然后介绍表格，我们的3种model setting，我们评估了哪些metric。
%第二段描述结果，第二种setting比第一种好，说明ram什么什么的有效性。。。第三种比第二种好
The VizWiz dataset \cite{bigham2010vizwiz} is a collection of images taken by blind and individuals with visual disability, specifically designed to evaluate computer vision algorithms aimed at assisting individuals with visual disability. %It contains tasks of Visual Question Answering, Image Captioning, Image Classification and other more specific tasks for individuals with visual disability.
In this section, we conduct experiments on the task of Visual Question Answering \cite{gurari2018vizwiz} of VizWiz dataset to verify the effectiveness of our proposed method. 

We tested 4319 questions from the validation dataset, which can be categorized into four types, ``Unanswerable'', ``Other'', ``Yes/No'' and ``Number''. As for evaluation metrics, we use the BLEU \cite{papineni2002bleu}, ROUGE-L \cite{lin2004rouge}, METEOR \cite{banerjee2005meteor}, CIDEr \cite{vedantam2015cider} metrics to evaluate our results. Since the answer to Visual Question Answer is usually shorter than 4 words, we evaluate our results on BLEU\_1 and BLEU\_2. From Table \ref{tab:qresults}, our model averagely achieves quantitative results of 25.43 in BLEU\_1 and 53.98 in CIDEr, indicating the effectiveness \mbox{for pBLV.}

\begin{table}[H]
\caption{Quantitative %MDPI: moved the table after the first citation, please confirm.
 Results on Visual Question Answering. Column Q represents four question types and average results.}
\label{tab:qresults}

\newcolumntype{C}{>{\centering\arraybackslash}X}
\begin{tabularx}{\textwidth}{CCCCCC}
\toprule
\textbf{Q} & \textbf{BLEU\_1} & \textbf{BLEU\_2}  & \textbf{METEOR} & \textbf{ROUGE\_L} & \textbf{CIDEr} \\ \midrule
Unanswerable & 8.53 & 3.33  & 7.88 & 13.18 & 5.65 \\
Other & 34.27 & 20.75  & 23.22 & 44.95 & 74.72 \\
Yes/No & 50.94 & 9.10  & 38.24 & 73.90 & 112.82 \\
Number & 11.41 & 5.82  & 12.73 & 16.95 & 27.48 \\
Avg. & 25.43 & 14.52  & 19.09 & 35.76 & 53.98 \\ \bottomrule
\end{tabularx}

\end{table}

\subsection{Real-World Tests}

\begin{figure}[H] 
    
    \includegraphics[width=13.5cm]{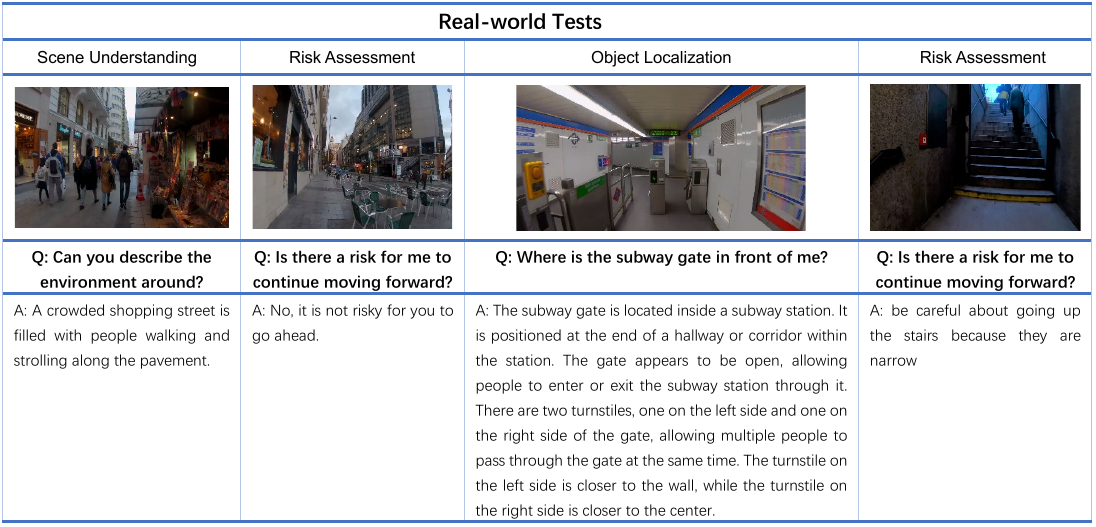}
    %\vspace{-0.5em}
    \caption{Examples of scene understanding, object detection and localization, and risk assessment under real-world settings.}
    %\vspace{-1.5em}
    \label{exp_2}
\end{figure}

We also conducted experiments to evaluate the proposed system in real-world situations as shown in Figure \ref{exp_2}. Specifically, we simulate the walking process of pBLV. The main content is a real video of a person walking on the street entering and then exiting a subway station in New York. Even though this route is quite simple in its geometry and layout, it may engender many risks for ppBLV. We captured several characteristic images from this video and passed them into our model for evaluation. These scenes are on the street, before entering the station, in the subway station, and after exiting the station.

The first scene shows a street with a crowd. Moreover, there is a shop on the left of the image. Our model returns the answer ``\textit{A crowded shopping street is filled with people walking and strolling along the pavement.}'', which is consistent with the image content. Our model identified the presence of shops on both sides and a large crowd, inferring that this is a bustling street scene. This highlights the importance of caution and slow navigation for individuals with visual disability amidst the bustling crowd.

For the second scene, the user is walking on a straight and empty street, when the user asks ``\textit{Is there a risk for me to continue moving forward?}'' The model answers that ``\textit{No, it is not risky for you to go ahead.}'', which is also in line with the actual situation. From the picture, we can see that the protagonist is walking on the sidewalk. There is mostly an open area on the sidewalk, with only tables and chairs on the right side. Therefore, the model determines that the road on this side can be walked, which means there is no risk.

In the third scene, when the user asks where the subway gate is, the model provides a very detailed explanation of the location of the subway gate with directional adjectives, such as front, back, left, and right. From the answer, it is clear that there are two gates and they are on the left and right sides. 

In the last scenario where there is a staircase, the model reminds the user that there currently exists a certain level of risk due to the presence of stairs. Therefore, the model provides this answer ``\textit{be careful about going up the stairs because they are narrow}''. This indicates that our model can effectively assess whether the current situation poses some risk to individuals with visual disability in moving forward.

\section{Conclusions}

In this paper, we present a pioneering approach that addresses the challenges faced by people with blindness and low vision (pBLV) in comprehensive scene understanding, precise object localization, and risk assessment in unfamiliar environments. By leveraging a large vision language model and integrating it with an image tagging module, our method provides pBLV with detailed and comprehensive descriptions and guidance to address their specific needs. We evaluate our approach through experiments conducted on both indoor and outdoor datasets. Our results demonstrate that our method can recognize objects accurately and provides insightful descriptions and analysis for pBLV.

Theoretical contributions of our work include a thorough exploration of the integration of vision language models with deep learning techniques for assisting individuals with visual impairments. On the other hand, practical contributions encompass the development of a functional system capable of providing detailed scene descriptions and guidance to enhance the mobility and independence of individuals with visual impairments. Enhancing the mobility of individuals with visual impairments is of paramount importance in fostering inclusivity and independence. These advancements will not only improve the practical utility of our system but also contribute to the theoretical understanding of assistive technologies for individuals with visual impairments, paving the way for more inclusive and accessible environments.

\section{Limitations and Future Research}

Despite these advances, we acknowledge certain limitations inherent to our system. The dynamic and complex nature of real-world environments poses a significant challenge. Factors such as changing lighting conditions, weather variations, and the presence of moving objects can impact the system's ability to accurately interpret scenes and predict potential risks, occasionally leading to false alarms. Moreover, the effectiveness of our system is contingent upon the quality of input data. Images captured by smartphone cameras under suboptimal conditions, such as poor lighting or obstructions, can adversely affect the model's performance, resulting in inaccurate recognitions or missed detections.

Inherent Limitations of AI Models: Despite the advanced capabilities of our Recognize Anything Model (RAM) and the InstructBLIP vision-language model, AI-based systems are not infallible. They operate within the confines of their training data and algorithms, which might not cover every possible real-world scenario or object encountered by people with blindness and low vision (pBLV). This limitation can lead to inaccuracies or false positives in object detection and scene interpretation.

Dynamic and Complex Environments: Real-world environments are highly dynamic and complex, with constant changes that can challenge the model's ability to accurately interpret and predict risks. Factors such as varying lighting conditions, weather changes, and moving objects can affect the system's performance and potentially lead to false alarms.

Quality of Input Data: The effectiveness of our system heavily relies on the quality of the input data, i.e., the images captured by the smartphone camera. Blurred images, poor lighting conditions, or obstructed views can hinder the model's ability to accurately recognize and analyze the scene, leading to potential false alarms or missed detections.

In future work, we aim to refine our approach by expanding training datasets to include a wider variety of environmental conditions, thereby enhancing the model's ability to generalize across different scenarios. We will also focus on developing more sophisticated algorithms that can more effectively distinguish between transient and permanent features in the environment, reducing false alarms. Additionally, integrating multimodal data sources, including auditory and haptic feedback, will be explored to compensate for visual data limitations and improve scene analysis. Incorporating user feedback mechanisms will further allow the system to adapt and learn from real-world applications, continually improving its performance and reliability for assisting individuals with blindness and \mbox{low vision. }

%%%%%%%%%%%%%%%%%%%%%%%%%%%%%%%%%%%%%%%%%%
% \section{Patents}

% This section is not mandatory, but may be added if there are patents resulting from the work reported in this manuscript.

% %%%%%%%%%%%%%%%%%%%%%%%%%%%%%%%%%%%%%%%%%%
% \vspace{6pt} 

%%%%%%%%%%%%%%%%%%%%%%%%%%%%%%%%%%%%%%%%%%
%% optional
%\supplementary{The following supporting information can be downloaded at:  \linksupplementary{s1}, Figure S1: title; Table S1: title; Video S1: title.}


%\begin{tabularx}{\textwidth}{lll}
%\toprule
%\textbf{Name} & \textbf{Type} & \textbf{Description} \\
%\midrule
%S1 & Python script (.py) & Script of python source code used in XX \\
%S2 & Text (.txt) & Script of modelling code used to make Figure X \\
%S3 & Text (.txt) & Raw data from experiment X \\
%S4 & Video (.mp4) & Video demonstrating the hardware in use \\
%\bottomrule
%\end{tabularx}
%}

%%%%%%%%%%%%%%%%%%%%%%%%%%%%%%%%%%%%%%%%%%
 \authorcontributions{Conceptualization, Y.H., F.Y, Y.F.; methodology, Y.H., F.Y, H.H., Y.W., Y.F.; software, Y.H., F.Y.; validation, Y.H., F.Y., H.H., S.Y., S.R., J.R., Y.W., Y.F.; writing---original draft preparation, Y.H., F.Y.; writing---review and editing, Y.H., F.Y., H.H., S.Y., S.R., J.R., Y.W., Y.F.; supervision, Y.F.; project administration, Y.F.; funding acquisition, J.R., Y.W., Y.F, S.R. All authors have read and agreed to the published version of the manuscript.} %MDPI: For research articles with several authors, a short paragraph specifying their individual contributions must be provided. The following statements should be used ``Conceptualization, X.X. and Y.Y.; methodology, X.X.; software, X.X.; validation, X.X., Y.Y. and Z.Z.; formal analysis, X.X.; investigation, X.X.; resources, X.X.; data curation, X.X.; writing---original draft preparation, X.X.; writing---review and editing, X.X.; visualization, X.X.; supervision, X.X.; project administration, X.X.; funding acquisition, Y.Y. All authors have read and agreed to the published version of the manuscript.'', please turn to the  \href{http://img.mdpi.org/data/contributor-role-instruction.pdf}{CRediT taxonomy} for the term explanation. Authorship must be limited to those who have contributed substantially to the work~reported.

 \funding{This study was supported by the National Science Foundation under Award Number ITE2345139, and number R33EY033689. The content is solely the responsibility of the authors and does not necessarily represent the official views of the National Institutes of Health and the National Science Foundation.} %MDPI: Please add: ``This research received no external funding'' or ``This research was funded by NAME OF FUNDER grant number XXX.'' and  and ``The APC was funded by XXX''. Check carefully that the details given are accurate and use the standard spelling of funding agency names at \url{https://search.crossref.org/funding}, any errors may affect your future funding.

 \institutionalreview{Not applicable.} %MDPI: In this section, you should add the Institutional Review Board Statement and approval number, if relevant to your study. You might choose to exclude this statement if the study did not require ethical approval. Please note that the Editorial Office might ask you for further information. Please add “The study was conducted in accordance with the Declaration of Helsinki, and approved by the Institutional Review Board (or Ethics Committee) of NAME OF INSTITUTE (protocol code XXX and date of approval).” for studies involving humans. OR “The animal study protocol was approved by the Institutional Review Board (or Ethics Committee) of NAME OF INSTITUTE (protocol code XXX and date of approval).” for studies involving animals. OR “Ethical review and approval were waived for this study due to REASON (please provide a detailed justification).” OR “Not applicable” for studies not involving humans or animals.

 \informedconsent{Not applicable.} %MDPI: Any research article describing a study involving humans should contain this statement. Please add ``Informed consent was obtained from all subjects involved in the study.'' OR ``Patient consent was waived due to REASON (please provide a detailed justification).'' OR ``Not applicable'' for studies not involving humans. You might also choose to exclude this statement if the study did not involve humans.

% Written informed consent for publication must be obtained from participating patients who can be identified (including by the patients themselves). Please state ``Written informed consent has been obtained from the patient(s) to publish this paper'' if applicable.

 \dataavailability{Not applicable.} %MDPI: We encourage all authors of articles published in MDPI journals to share their research data. In this section, please provide details regarding where data supporting reported results can be found, including links to publicly archived datasets analyzed or generated during the study. Where no new data were created, or where data is unavailable due to privacy or ethical restrictions, a statement is still required. Suggested Data Availability Statements are available in section ``MDPI Research Data Policies'' at \url{https://www.mdpi.com/ethics}.

% % Only for journal Nursing Reports
% %\publicinvolvement{Please describe how the public (patients, consumers, carers) were involved in the research. Consider reporting against the GRIPP2 (Guidance for Reporting Involvement of Patients and the Public) checklist. If the public were not involved in any aspect of the research add: ``No public involvement in any aspect of this research''.}

% % Only for journal Nursing Reports
% %\guidelinesstandards{Please add a statement indicating which reporting guideline was used when drafting the report. For example, ``This manuscript was drafted against the XXX (the full name of reporting guidelines and citation) for XXX (type of research) research''. A complete list of reporting guidelines can be accessed via the equator network: \url{https://www.equator-network.org/}.}

% \acknowledgments{In this section you can acknowledge any support given which is not covered by the author contribution or funding sections. This may include administrative and technical support, or donations in kind (e.g., materials used for experiments).}

 \conflictsofinterest{The authors declare no conflict of interest.} %MDPI: Declare conflicts of interest or state ``The authors declare no conflict of interest.'' Authors must identify and declare any personal circumstances or interest that may be perceived as inappropriately influencing the representation or interpretation of reported research results. Any role of the funders in the design of the study; in the collection, analyses or interpretation of data; in the writing of the manuscript; or in the decision to publish the results must be declared in this section. If there is no role, please state ``The funders had no role in the design of the study; in the collection, analyses, or interpretation of data; in the writing of the manuscript; or in the decision to publish the results''.

\begin{adjustwidth}{-\extralength}{0cm}
%\centering %% If there is a figure in wide page, please release command \centering

\reftitle{References}

\PublishersNote{}

\end{adjustwidth}
\end{document}